\documentclass[11pt]{article}

\usepackage{amsmath}
\usepackage{amssymb}
\usepackage{amsthm}
\usepackage{graphicx}
\usepackage{url}
\usepackage{hyperref}
\usepackage{algorithm}
\usepackage{algpseudocode}
\usepackage{tikz}
\usepackage{arydshln}
\usepackage{natbib}
\usetikzlibrary{arrows.meta, positioning, calc, shapes.multipart, decorations.pathmorphing}

\usepackage[margin=1in]{geometry}
\usepackage{times}
\usepackage{parskip}

\theoremstyle{plain}
\newtheorem{theorem}{Theorem}[section]

\theoremstyle{definition}
\newtheorem{definition}[theorem]{Definition}

\newcommand{\method}{GLIDR}

\title{GLIDR: Graph-Like Inductive Logic Programming with Differentiable Reasoning}

\author{
Blair Johnson \\
Georgia Institute of Technology \\
\texttt{blair.johnson@gtri.gatech.edu} 
\and
Clayton Kerce \\
Georgia Institute of Technology \\
\texttt{clayton.kerce@gtri.gatech.edu}
\and
Faramarz Fekri \\
Georgia Institute of Technology \\
\texttt{ffekri@ece.gatech.edu}
}

\date{}

\begin{document}

\maketitle

\begin{abstract}
Differentiable inductive logic programming (ILP) techniques have proven effective at finding approximate rule-based solutions to link prediction and node classification problems on knowledge graphs; however, the common assumption of chain-like rule structure can hamper the performance and interpretability of existing approaches.
We introduce GLIDR, a differentiable rule learning method that models the inference of logic rules with more expressive syntax than previous methods. GLIDR uses a differentiable message passing inference algorithm that generalizes previous chain-like rule learning methods to allow rules with features like branches and cycles. GLIDR has a simple and expressive rule search space which is parameterized by a limit on the maximum number of free variables that may be included in a rule. Explicit logic rules can be extracted from the weights of a GLIDR model for use with symbolic solvers.
We demonstrate that GLIDR can significantly outperform existing rule learning methods on knowledge graph completion tasks and even compete with embedding methods despite the inherent disadvantage of being a structure-only prediction method. We show that rules extracted from GLIDR retain significant predictive performance, and that GLIDR is highly robust to training data noise. Finally, we demonstrate that GLIDR can be chained with deep neural networks and optimized end-to-end for rule learning on arbitrary data modalities.
\end{abstract}

% Include sections
\section{Introduction}
This paper presents GLIDR (Graph-like Logical Induction with Differentiable Reasoning), a differentiable inductive logic programming (ILP) method that learns expressive graph-structured logical rules for knowledge graph reasoning. Unlike existing differentiable ILP approaches that are limited to chain-like rule structures, GLIDR supports an expressive syntax that can represent rules with branches, cycles, and complex variable interactions. The method consists of two key components: (1) a graph-like rule representational structure, and (2) a differentiable message passing algorithm for deriving the entailment of rules encoded in this representational structure. Experiments on knowledge graph completion benchmarks demonstrate GLIDR's expanded expressiveness enables it to significantly outperform existing rule learning methods while maintaining the noise robustness and scalability advantages of differentiable approaches. Furthermore, as a differentiable method, GLIDR can be integrated end-to-end with deep neural networks, enabling rule learning on mixed symbolic and continuous data modalities.

\subsection{Inductive Logic Programming}
Inductive Logic Programming (ILP) is a machine learning technique that learns from examples of facts to construct logical rules. ILP benefits from a strong modeling bias that allows it to learn from small amounts of data, and the logical nature of the models it produces is explicitly interpretable. Learned rules explicitly describe the conditions under which a model makes a given prediction. In doing this, learned rules often reveal explicit knowledge about the patterns present in the data used for training.

The logical rules learned by ILP systems are typically first-order logic rules comprised of predicates, variables, constants, and logical operators.
A predicate $P$ can represent conceptual relationships or properties. Constants represent specific entities, and predicates can be combined with constants to represent facts. For instance, $\text{dog}(\text{Rufus})$, uses the "dog" predicate to record the fact that a specific entity, "Rufus", is a dog.
Sometimes, we need to express facts that are true for many entities without enumerating them.
Logic variables allow first-order logic to express abstract facts in this manner.
The fact $\forall X,Y(\text{parent}(X)\leftarrow \text{has\_child}(X,Y))$ states that for all possible pairs of entities, represented by variables $X,Y$, if the entity represented by $Y$ is the child of the entity represented by $X$, then $X$ must necessarily be a parent.
We call abstract facts of this nature rules because they describe patterns of other facts.
The first term of the rule, $\text{parent}(X)$, is referred to as the \emph{head literal} and is implied to be true whenever the fact described by the \emph{body literal}, $\text{has\_child}(X,Y)$, is true.
Logical operators like conjunction ($\wedge$) and disjunction ($\vee$) allow first order logic to express more complex facts.
The conjunction operator represents the logical AND of facts, and it is frequently used to combine multiple literals in a rule body to express simultaneously necessary conditions. 
For example, the rule $\forall X,Y,Z($grandparent$(X,Y)\leftarrow$parent$(X,Z)\wedge$parent$(Z,Y))$ represents the fact that an entity $X$ is always the grandparent of an entity $Y$ if $X$ is the parent of an entity $Z$ \emph{and} that entity $Z$ is the parent of $Y$.
The disjunction operator implements the logical OR of facts, meaning that at least one of the facts it connects must be true.
ILP systems typically favor conjunctive rule bodies because they are computationally efficient to evaluate and describe clear, human-understandable sets of truth conditions.
Using these simple primitives, rules learned by ILP can represent highly expressive statements about the world.

\subsection{Knowledge Graphs}
As ILP deals heavily with relational facts, ILP problems are closely related to knowledge graph reasoning problems.
Knowledge graphs are useful and convenient data structures for encoding information about entities, their properties, and the relationships between them.
The generality of knowledge graphs as a mechanism for storing and retrieving relational information has resulted in widespread adoption across many domains such as biomedicine \cite{nicholson2020constructing, ernst2015knowlife}, cybersecurity \cite{piplai2020cyber, qi2023attack}, financial crime \cite{suzumura2019towards}, and public health \cite{park2021discovering, bettencourt2020exploring}.
By representing knowledge in a structured and machine-readable format, knowledge graphs have enabled a wide range of applications, including question answering, recommendation systems \cite{Fensel2020}, and decision support systems \cite{nickel2015review}. All of these applications rely on the core tasks of predicting edges and classifying nodes using the background knowledge contained in the knowledge graph. These tasks can also be represented as ILP problems, and ILP methods are frequently evaluated on knowledge graph completion benchmarks.

While embedding methods represent the most popular and performant approaches for knowledge graph reasoning tasks, they have several important limitations. The most significant limitation of embedding methods is that they typically operate in the \emph{transductive} setting, and must be trained on a set of entities that overlaps those seen at test-time \cite{teru2020inductive}. They often perform poorly on rare entities with few training examples \cite{zhang2019iteratively}. They have limited interpretability \cite{GeseseBS19, HamiltonYL17}, and they struggle to model patterns that can be naturally compressed into rules \cite{qu2020rnnlogic}. These limitations motivate the development of graph reasoning algorithms that can generate predictions from \emph{structure alone}, enabling interpretability and generalization unseen data.

\subsection{Previous Work}
Previous ILP techniques such as Progol \cite{muggleton1995progol}, Aelph \cite{srinivasan_aleph}, and Metagol \cite{muggleton_meta-interpretive_2015} have relied heavily on symbolic solvers, often implemented in Prolog, to search for rules that entailed the positive examples provided to the algorithm and reject the negative examples. These solvers used hypothesis generation algorithms, inverse entailment, heuristic search, and meta-templating strategies to construct rules explaining observed data.
Recent advancements such as Popper \cite{cropper_learning_2021} have re-framed ILP problems as special cases of answer set programming (ASP) or Boolean satisfiability (SAT) problems. These approaches leverage high-performance SAT and ASP solvers to improve the performance and efficiency of the ILP process while enjoying the passive benefit of advancements in SAT and ASP solving.
While symbolic ILP techniques like Metagol and Popper initially lacked noise tolerance, extensions to these methods have eased this limitation \cite{muggleton2018meta, noisypopper, hocquette_learning_2023}. 

Today's symbolic ILP methods are very capable but have properties that make them unsuitable for some problems. First, the computational efficiency of these systems is often extremely dependent on the choice of language bias supplied to the solver, and choosing such a bias can be a difficult process requiring both ILP and problem domain knowledge \cite{cropper2022inductive}. Second, symbolic ILP methods often struggle when very large amounts of noisy background knowledge is supplied during rule learning. Finally, symbolic ILP methods do not interface with continuous data well, as they typically operate entirely on explicit symbolic representations of data \cite{cropper2022inductive}. Some recent works, such as \cite{prob_popper} have begun to address this limitation for probabilistic data.

Recent work on differentiable ILP solvers has attempted to address some of these issues by employing gradient-based optimization methods to search for rules in a numerically parameterized search space \cite{cohen2016tensorlog, yang2017differentiable, evans2018learning, sadeghian2019drum,  payani2019inductive, yang2020learn}. These differentiable methods are noise robust, and some can scale to datasets with very large numbers of facts and predicate types. The numerical nature of these methods has also allowed some to interface with continuous valued data or even other machine learning models for end-to-end learning on mixed symbolic and non-symbolic domains \cite{evans2018learning}. 

While these are valuable properties, most differentiable ILP methods cannot represent or learn the full scope of language features supported by some symbolic methods such as N-ary predicates, lists, number systems, and recursive rule evaluation. These methods also vary significantly in their scalability and expressiveness. Methods such as $\delta$ILP \citep{evans2018learning} and dNL-ILP \citep{payani2019inductive} can represent complex rules supporting more traditional language features, but this comes at a significant memory cost that seriously restricts the size of problems that they can be applied to. Other, more memory efficient ILP methods such as Neural-LP \cite{yang2017differentiable}, DRUM \cite{sadeghian2019drum}, and NLIL \cite{yang2020learn} can scale to much larger datasets, but they adopt restrictive rule syntax and support very few language features.
\method\ improves on the inference algorithm used by these methods to enable differentiable learning of rules that have more general syntax. In doing so, \method\ can learn relationships that are impossible to represent for previous memory-efficient differentiable ILP methods. In this paper we demonstrate that:
\begin{enumerate}
    \item The added expressiveness of our method translates into performance improvements on knowledge graph completion tasks.
    \item Our method retains performance when subject to noisy training data.
    \item Hard rules extracted from our method retain significant performance.
    \item Our method can be effectively co-trained with other deep learning models using end-to-end optimization.
\end{enumerate}

\subsection{Chain-Like Rule Syntax}
Most differentiable ILP methods can learn rules that conform to a schema that we describe as "chain-like".

\begin{definition} \label{def:chainlike}
Any $n$-variable logic rule that can be equivalently expressed in the standard form described by Equation \ref{eq:chainlike} is chain-like.
\begin{equation} \label{eq:chainlike}
    \forall \{Z_i\}_{i=1}^n\left(P_h(Z_1,Z_n)\leftarrow \bigwedge_{i=1}^{n-1} P_{i}(Z_{i},Z_{i+1})\right)
\end{equation}
\end{definition}

In this schema, there are $n$ logic variables with $n-1$ binary (2-ary) body literals that form a "chain" between the head variables. 
In practice, the restriction that argument order must follow the $P({Z_i,Z_{i+1}})$ pattern is typically relaxed by introducing "inverse" predicates such that $P_{inv}(X,Y) \Leftrightarrow P(Y,X).$
This means that both $P_{head}(X,Y)\leftarrow P_A(X,Z_1),P_B(Z_1,Z_2),P_C(Z_2,Y)$ and $P_{head}(X,Y)\leftarrow P_A(X,Z_1),P_B(Z_2,Z_1),P_C(Y,Z_2)$ can be made into valid chain-like rules. 

Chain-like rules are favored by most differentiable ILP methods because rule inference is computationally convenient.
Chain-like rules are also very natural for answering queries that commonly appear in information retrieval problems. For instance, the query "Who is the brother of the director of the most recent Star Wars movie?" can be easily modeled as a chain of relations:
\begin{equation}
    ?-\text{brother}(X,Z_1),\text{directedMovie}(Z_1,Z_2),\text{mostRecentEntry}(Z_2,Z_3),\text{StarWarsFranchise}(Z_3).
\end{equation}

GLIDR extends previous differentiable rule learning methods with a more complex rule inference algorithm based on message passing. 
This updated algorithm enables GLIDR to perform inference on non-chain-like, or what we call "graph-like" rules.
We will show that any rule conforming to the chain-like schema can also be expressed using our graph-like schema, demonstrating that the latter is strictly more general.

\section{Methodology}
\newcommand{\child}[1]{\text{Ch}(#1)}
\newcommand{\parent}[1]{\text{Pa}(#1)}
\newcommand{\gadj}{\mathbf{B}}
\newcommand{\satvar}[1]{\phi^{(#1)}}
\newcommand{\head}{e_x}
\newcommand{\tail}{e_y}
\newcommand{\bkg}{\mathbf{B}}

GLIDR consists of (1) a graph-like rule representational structure, and (2) a differentiable message passing algorithm for deriving the entailment of rules encoded in this representational structure. These two features allow GLIDR to be optimized using gradient-based optimization to learn rules that conform to its representational structure and predict training data.

\subsection{Overview of GLIDR's Operational Modes}
GLIDR's differentiable design gives rise to two distinct operational modes primarily distinguished by how predicate selections are represented in the model weights.
\begin{itemize}
\item \textbf{Soft Setting:} In the soft setting, the contributions of predicates to the modeled logic rule are modulated by confidence values from discrete probability distributions formed from the model's weights. During learning, candidate predicates are represented as linear superpositions of all possible predicates, and the learning process collapses these to predicate instances that are most consistent with the data. 
Consequentially, GLIDR's inference results in this setting are a "soft" \emph{approximation}, blending influences from multiple logical paths and generally not guaranteed to match those of any particular rule.
While not necessarily obeying strictly logical inference, this mode can still offer a degree of interpretability, as the learned confidence values may highlight dominant predicate choices and emergent rule-like structures.

\item \textbf{Hard Setting:} In the hard setting, GLIDR uses strictly binary valued weights encoding a specific logic rule.
In this setting, GLIDR's inference can exactly model logical rule inference for certain classes of rules and approximately model inference for others. Section \ref{sec:theory} discusses the distinction. The ability to differentiably approximate inference on specific "hard" rules can be useful in special applications where a differentiable rule inference algorithm is required. 
When differentiability is not required, symbolic solvers offer more efficient and exact rule inference.
\end{itemize}
GLIDR can be used to identify logical rules by applying a \emph{rule extraction algorithm} to the model weights learned in the soft setting.
Rules extracted from GLIDR's weights are strictly logical, offering direct human interpretability.
Generally, there is some task performance lost when extracting a logical rule from GLIDR, however this depends on the degree of convergence in the model weights. It is possible for GLIDR to converge to weight configurations in the soft setting that correspond closely to discrete logic rules, enabling high-fidelity rule extraction.

\subsection{GLIDR's Graph-Like Rule Syntax} \label{sec:graph-like-syntax}
The key innovation that allows GLIDR to outperform other differentiable rule learning methods is support for a more general class of logic rules. 
GLIDR can learn rules which conform to a more expressive schema than chain-like rules. We describe rules that conform to this schema as "graph-like".

\begin{figure}[htbp]
    \centering
    \begin{tikzpicture}
      \node (n1) at (0,0) [circle, draw, minimum size=.5cm] {$Z_{1}$};
      \node (n2) at (2.5,0) [circle, draw, minimum size=.5cm] {$Z_2$};
      \node (n3) at (5,0) [circle, draw, minimum size=.5cm] {$Z_3$};
      \node (n4) at (7.5,0) [circle, draw, minimum size=.5cm] {$Z_4$};
    
      \draw[-{Stealth}, bend left=0] (n1) to node[above] {$P_{1,2}$} (n2);
      \draw[-{Stealth}, bend left=40] (n1) to node[above] {$P_{1,3}$} (n3);
      \draw[-{Stealth}, bend left=60] (n1) to node[above] {$P_{1,4}$} (n4);
    
      \draw[-{Stealth}, bend left=0] (n2) to node[above] {$P_{2,3}$} (n3);
      \draw[-{Stealth}, bend right=40] (n2) to node[above] {$P_{2,4}$} (n4);
    
      \draw[-{Stealth}, bend left=0] (n3) to node[above] {$P_{3,4}$} (n4);
    \end{tikzpicture}
    \caption{Graphical illustration of the variables and predicates in the graph-like rule schema for $n=4$. The graph-like schema invokes a directed acyclic graph (DAG) representation.}
    \label{fig:rule-graph}
\end{figure}
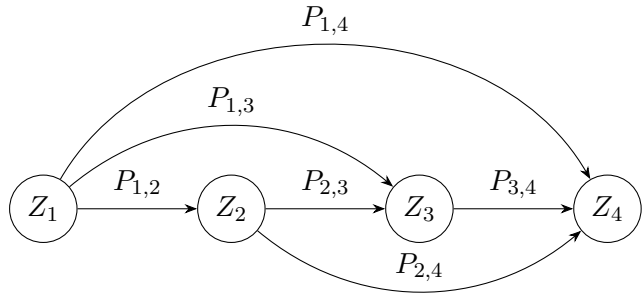

\begin{definition} \label{def:graphlike}
Any $n_v$-variable logic rule that can be equivalently expressed in the standard form described by Equation \ref{eq:graphlike} (with $n\geq n_v$ variables) is a graph-like rule.
\begin{equation} \label{eq:graphlike}
    \forall \{Z_i\}_{i=1}^n\left(P_h(Z_1,Z_n)\leftarrow \bigwedge_{i=1}^{n-1} \bigwedge_{j=i+1}^nP_{i,j}(Z_i,Z_j)\right)
\end{equation}
\end{definition}
Equation \ref{eq:graphlike} defines the graph-like rule schema involving $n$ distinct variables, $Z_1, \ldots, Z_n$. The head of the rule, $P_h\left(Z_1, Z_n\right)$, uses the first and last variables of this indexed set. The body is a conjunction of $n(n-1) / 2$ binary literals, $P_{i, j}\left(Z_i, Z_j\right)$, one for each distinct pair of variables $\left(Z_i, Z_j\right)$ where $i<j$. This schema defines a maximal set of potential directed interactions between variables. Figure \ref{fig:rule-graph} illustrates this maximal, directed acyclic graph (DAG) of literals for $n=4$ variables.

\subsection{Equivalently Expressing Graph-Like Rules} \label{sec:extended_bg_predicates}
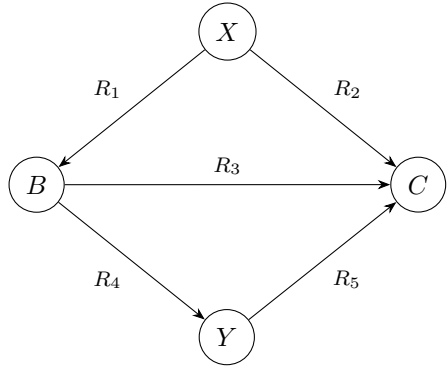
\begin{figure}[htbp]
    \centering
    \begin{tikzpicture}[
        node distance=1.5cm and 2cm,
        mynode/.style={circle, draw, minimum size=0.7cm, font=\small}
    ]
        \node[mynode] (A) {$X$};
        \node[mynode, below left=of A] (B) {$B$};
        \node[mynode, below right=of A] (C) {$C$};
        \node[mynode, below right=of B] (D) {$Y$};

        \draw[-{Stealth}] (A) to node[above left, midway, font=\scriptsize] {$R_1$} (B);
        \draw[-{Stealth}] (A) to node[above right, midway, font=\scriptsize] {$R_2$} (C);
        \draw[-{Stealth}] (B) to node[above, midway, font=\scriptsize] {$R_3$} (C);
        \draw[-{Stealth}] (B) to node[below left, midway, font=\scriptsize] {$R_4$} (D);
        \draw[-{Stealth}] (D) to node[below right, midway, font=\scriptsize] {$R_5$} (C);
    \end{tikzpicture}
    \caption{Conceptual graph of the example rule $P_h(X,Y) \leftarrow R_1(X,B) \land R_2(X,C) \land R_3(B,C) \land R_4(B,Y) \land R_5(Y,C)$.}
    \label{fig:conceptual_rule}
\end{figure}

The implicit assertion in Definition \ref{def:graphlike} that a wide range of $n_v$-variable rules can be "equivalently expressed" by the graph-like schema hinges on the flexibility afforded by freely selecting predicates for each $P_{i,j}$ "slot". 
When provided with a set of background predicates, $\mathcal{P}_{bg}$, with which to build a rule, GLIDR automatically constructs an extended set of background predicates, $\mathcal{P}^*_{bg}$, which includes:

\begin{itemize}

    \item \textbf{Standard Domain Predicates:} The set of available domain predicates from $\mathcal{P}_{bg}$. These comprise the union of all predicates represented in the background knowledge.

    \item \textbf{A "Null" Predicate ($P_{true}$):} This predicate is universally true (e.g., $P_{true}(A,B)$ holds for any $A,B$). When a schema slot $P_{i,j}$ is instantiated with $P_{true}$, the literal $P_{true}(Z_i,Z_j)$ becomes logically redundant within the body's conjunction. 
    This effectively "masks" or removes that specific $P_{i,j}(Z_i,Z_j)$ literal from the rule body. This predicate allows the maximal schema to represent rules with sparser graphical structures.

    \item \textbf{Inverse Predicates ($P_{inv}$):} For every predicate $P(X,Y)$ from the domain $\mathcal{P}_{bg}$, the inverse predicate $P_{inv}(X,Y)\Leftrightarrow P(Y,X)$ is included in $\mathcal{P}^*_{bg}$.
    As in the case of chain-like rules, this provides flexibility in constructing literals with argument orderings that disobey the schema's fixed $Z_i \to Z_j$ slot structure.

\end{itemize}
The use of these additional background predicates allows GLIDR to represent an expansive and diverse set of logic rules. To illustrate this, consider the logic rule depicted in Figure \ref{fig:conceptual_rule}, $P_h(X,Y) \leftarrow R_1(X,B) \land R_2(X,C) \land R_3(B,C) \land R_4(B,Y) \land R_5(Y,C)$. This rule is distinctly non-chain-like because the intermediate variables $B,C$ each appear in more than 2 literals, meaning that they must "branch" and violate the multi-hop structure of chain-like rules.

\begin{figure}
    \centering
    \begin{tikzpicture}[
        node distance=1.6cm,
        mynode/.style={circle, draw, minimum size=0.7cm, font=\small},
        active_edge/.style={-{Stealth}, draw=black, thick},
        ptrue_edge/.style={-{Stealth}, draw=black, densely dotted, thin}
    ]
        \node[mynode] (Z1) {$Z_1$};
        \node[mynode, right=of Z1] (Z2) {$Z_2$};
        \node[mynode, right=of Z2] (Z3) {$Z_3$};
        \node[mynode, right=of Z3] (Z4) {$Z_4$};
        \node[mynode, right=of Z4] (Z5) {$Z_5$};

        \draw[active_edge] (Z1) to node[above, font=\scriptsize, midway] {$R_1$} (Z2); 
        \draw[active_edge, bend left=30] (Z1) to node[above, font=\scriptsize, midway] {$R_2$} (Z3); 
        \draw[active_edge] (Z2) to node[above, font=\scriptsize, midway] {$R_3$} (Z3); 
        \draw[active_edge, bend right=40] (Z2) to node[above, font=\scriptsize, midway] {$R_4$} (Z5); 
        \draw[active_edge, bend right=20] (Z3) to node[below, font=\scriptsize, midway] {$R_{5,inv}$} (Z5); 

        \draw[ptrue_edge, bend left=40] (Z1) to (Z4);
        \draw[ptrue_edge, bend left=50] (Z1) to (Z5);
        \draw[ptrue_edge, bend left=30] (Z2) to (Z4);
        \draw[ptrue_edge] (Z3) to (Z4);
        \draw[ptrue_edge] (Z4) to (Z5);
    \end{tikzpicture}
    \caption{Example rule from Figure \ref{fig:conceptual_rule} equivalently represented within GLIDR's graph-like schema for $n=5$ variables ($Z_1 \equiv X, Z_2 \equiv B, Z_3 \equiv C, Z_5 \equiv Y$; $Z_4$ is unused). Solid lines are active predicates from the rule (e.g., $P_{1,2} \equiv R_1$, $P_{3,5} \equiv R_{5,inv}$); dotted lines represent schema slots occupied by $P_{true}$.}
    \label{fig:glidr_schema_instance}
\end{figure}
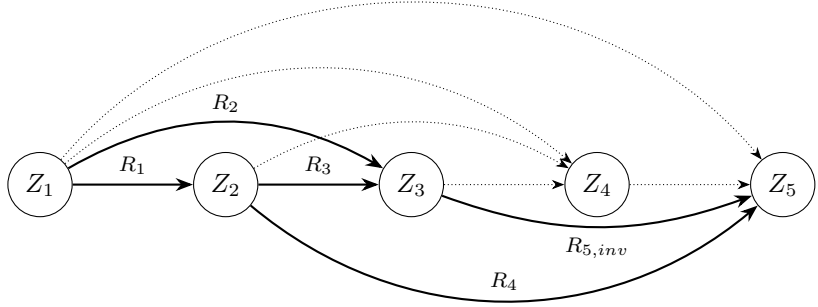

Figure \ref{fig:glidr_schema_instance} illustrates how this 4-variable conceptual rule is equivalently expressed using GLIDR's graph-like schema, instantiated here with $n=5$ schematic variables ($Z_1,\dots,Z_5$). 
This representation is achieved by first mapping the conceptual rule's four variables $(X,B,C,Y)$ to a subset of the five schema variables $(Z_1,\dots,Z_5)$. This leaves one schema variable $(Z_4)$ effectively unused in this particular rule. Then, the appropriate $P_{i,j}$ predicates in the schema are instantiated with the conceptual rule's predicates ($R_1,\dots,R_5$), utilizing an inverse predicate ($R_{5,inv}$) to make $P_{3,5}$ represent $R_5(Y,C)$. Finally, the $P_{true}$ predicate is assigned to all remaining $P_{i,j}$ in the maximal rule (illustrated by dotted lines in the diagram).
This example highlights how an appropriate selection of each $P_{i,j}$ from $\mathcal{P}^*_{bg}$ allows GLIDR to precisely represent diverse graph-like rules.

\subsection{Rule Inference Characteristics}
GLIDR can learn rules that are syntactically recursive, meaning that the predicate in the rule head also appears in the rule body. GLIDR does not support recursive execution or unfolding of rules; all body literals are exclusively resolved by direct lookup against the ground facts in a background knowledge base. This means that rules learned by GLIDR can capture recursive patterns but evaluation of such rules is limited to single-step deductive lookup.
Similarly, GLIDR can not use logic rules as background knowledge without first deriving their consequences as ground facts. All background knowledge used by GLIDR must be expressed as ground facts.

GLIDR's inference algorithm performs ground query evaluation and does not generally support open queries (queries with ungrounded head variables). To evaluate queries of this form, GLIDR uses a generate and test approach, where all possible groundings of an ungrounded head variable are instantiated and tested.
This limitation is a consequence the more complex rule inference algorithm required to support graph-like rules, and it is a notable departure from previous works like Neural-LP, DRUM, and NLIL. 
The requirement of exhaustive constant enumeration means that rule inference with GLIDR does not scale well when addressing open queries on large knowledge bases. For example, GLIDR took around 40 hours to generate all tail predictions for the knowledge graph completion benchmark on the Freebase15k-237 dataset using 4 NVIDIA A100 GPUs.

Unlike previous methods such as Neural-LP, DRUM, and NLIL, the rule parameterization in GLIDR is fixed rather than generated from a neural network. This means that rules learned by GLIDR must be individually instantiated and trained for different target predicates. 
This is not a fundamental limitation of the method, and future work could generate GLIDR's weights with a neural network for added training efficiency.

\subsection{Differentiable Rule Inference via Iterative Consistency Propagation} \label{sec:message-passing}
The core inference mechanism in GLIDR, while implemented with differentiable linear algebra for end-to-end learning, conceptually mirrors the iterative enforcement of local constraints employed by arc consistency algorithms (e.g., AC-3 \cite{mackworth_ac3}) in constraint satisfaction problems (CSPs). 
During inference, GLIDR maintains a "soft" domain of potential entity groundings for each logic variable. Predicates in each body literal act as differentiable constraints acting on these variables.
The inference algorithm iteratively refines the soft domains for each variable via message passing until a fixed point is reached or unsatisfiability is detected.

\subsubsection{GLIDR Initialization and Data Inputs}
During inference and training, GLIDR operates on a background knowledge graph, denoted $\mathcal{G}_{bg}$, which contains all of the ground facts from which GLIDR will reason. Each ground fact is a triple ($(e_i,r_k,e_j)\equiv r_k(e_i,e_j).$) describing a relationship $r_k$ (the predicate with index $k$) between two entities $e_i,e_j$. GLIDR treats these relationships as directional, meaning that generally $(e_i,r_k,e_j)\neq(e_j,r_k,e_i)$.

\begin{definition} \label{def:background_graph}
A knowledge graph $\mathcal{G}$ contains all triples from a set of facts $\mathcal{F}$. Each triple contains an ordered pair of entities from the set of graph entities, $e_i,e_j\in\mathcal{E}$, and a relation $r\in\mathcal{P}$ from the set of graph predicates. A unique index is defined on the sets of entities and relations such that $(e_i=e_j)\Leftrightarrow (i=j)$ and $(r_k=r_l)\Leftrightarrow (k=l)$.
    \begin{equation} \label{eq:background_graph}
        \mathcal{G}=\{(e_i,r_k,e_j)_{l}\}_{l=1}^{|\mathcal{F}|}
    \end{equation}
\end{definition}
GLIDR encodes the background graph $\mathcal{G}_{bg}$ into a numerical representation for inference. First, the set of background predicates is augmented with the inclusion of inverse predicates and $P_{true}$ to create the extended background predicate set $\mathcal{P}_{bg}^*$ as described in Section \ref{sec:extended_bg_predicates}. Next, $\mathcal{F}_{bg}$ is augmented with the addition of ground facts for each inverse predicate, such that:
\begin{equation}
\mathcal{F}_{bg}^*=\mathcal{F}_{bg}\cup\{(e_j,r_{k,inv},e_i):\forall(e_i,r_k,e_j)\in\mathcal{F}_{bg}\}
\end{equation}
Finally, we construct a sparse binary adjacency tensor $\gadj\in\{0,1\}^{|\mathcal{E}_{bg}|\times|\mathcal{E}_{bg}|\times|\mathcal{P}^*_{bg}|}$. 
Each element of this adjacency tensor encodes a triple from $\mathcal{F}^*_{bg}$ such that $((e_i,r_k,e_j)\in\mathcal{F}^*_{bg})\Leftrightarrow(\gadj_{j,i,k}=1)$. The adjacency matrix for $P_{true}$ is ill-defined and non-sparse, so this relationship is handled algorithmically and its corresponding slice of $\gadj$ is all zeros.
During inference, a GLIDR model is presented with a batch of triples $\{(e_i,r_k,e_j)\}_l$, with each $(e_i,e_j)$ encoded as one-hot vectors $x_i,x_j\in\{0,1\}^{|\mathcal{E}_{bg}|}:x_{i,i}=1,x_{j,j}=1$, and the background graph $\mathcal{G}_{bg}$ in the form of $\gadj$. For each triple, GLIDR produces a score $\hat{y}\in(0,1)$ indicating the likelihood of that fact being implied by the modeled rule for $r_k$ given $\mathcal{G}_{bg}$.

While the adjacency tensor $\gadj$ provides a fixed numerical encoding of the background knowledge, the specific rule structure that GLIDR models is determined by a set of learnable parameters, or weights. A GLIDR model is configured to represent any graph-like rule with $n_v\leq N$ distinct logic variables. It accomplishes this by initializing a numerical encoding of a graph-like rule with $n=N$ schematic variables following Equation \ref{eq:graphlike}.
For each of the $N(N-1)/2$ body literals in the schematic rule, GLIDR stores a learnable logit that encodes which predicate from the extended background set $\mathcal{P}_{bg}^*$ should occupy that slot in the rule body.
\begin{definition} \label{def:glidr_weights}
    For each predicate slot $(i,j):1\leq i < j \leq N$ in the $N$-variable graph-like schema defined in Equation \ref{eq:graphlike}, GLIDR maintains a vector of learnable logit parameters:
    \begin{equation} \label{eq:glidr_weights}
        \theta_{i,j}\in\mathbb{R}^{|\mathcal{P}_{bg}^*|}
    \end{equation}
    When the softargmax function $\sigma(\cdot)$ is applied to a vector of logits, the result is a discrete probability distribution over the set of predicates in $\mathcal{P}_{bg}^*$. The resulting probability weights $w_{i,j,k}$ encode the strength of predicate $r_k$'s contribution to the modeled body literal $P(Z_i,Z_j)$:
    \begin{equation}\label{eq:glidr_prob_weights}
        w_{i,j}\in\mathbb{R}^{|\mathcal{P}_{bg}^*|} = \sigma(\theta_{i,j})=\frac{e^{\theta_{i,j}}}{\sum_{k=1}^{|\mathcal{P}_{bg}^*|}e^{\theta_{i,j,k}}}
    \end{equation}
\end{definition}

\subsubsection{Numerical Representation of Variables and Predicates}
During inference, each logic variable (instantiated from the graph-like schema in Equation \ref{eq:graphlike} with $N$ nodes) $\{Z_i\}_{i=1}^N$ is associated with a \emph{state vector} $\phi_{i}\in\mathbb{R}^{|\mathcal{E}_{bg}|}$. Each element in a state vector $\phi_{i,l}\in[0,1]$ represents the current belief that the entity $e_l\in\mathcal{E}_{bg}$ satisfies all adjacent constraints imposed on the grounding of $Z_i$ by the body literals it is involved in. This is analogous to the domain $D(Z_i)$ tracked during an arc-consistency algorithm like AC-3 \cite{mackworth_ac3}. If at any time, $\max(\phi_i)=0$, then a domain wipeout has occurred, meaning that there are no possible groundings of $Z_i$ that satisfy the constraints placed upon it, and the rule must be false. At time step $t=0$ of inference, the first and last state vectors are initialized to equal the one-hot encodings of the input head entities $(e_i,e_j)$, and all other state vectors are initialized to $1$, indicating an unconstrained initial domain:
\begin{equation}
    \phi_k^{(0)} =
    \begin{cases}
      x_i & \text{if } k=1 \\
      x_j & \text{if } k = N \\
      \mathbf{1}^{|\mathcal{E}_{bg}|} & \text{else}
    \end{cases}
\end{equation}

Each predicate slot in the modeled rule schema $P_{i,j}(Z_i,Z_j)$ is modeled by a \emph{soft adjacency matrix} $\mathbf{M}_{i,j}\in\mathbb{R}^{|\mathcal{E}_{bg}|\times|\mathcal{E}_{bg}|}$ that acts as a learnable, differentiable constraint on the states $\phi_i,\phi_j$ of variables $Z_i,Z_j$.  Candidate predicates are represented as linear superpositions of all possible predicates (Equation \ref{eq:soft_adj_mtx}), and the learning process collapses this superposition to predicate instances that are most consistent with the data.
\begin{definition} The \emph{soft adjacency matrix} $\mathbf{M}_{i,j}\in\mathbb{R}^{|\mathcal{E}_{bg}|\times|\mathcal{E}_{bg}|}$ for graph-like rule schema slot $P_{i,j}(Z_i,Z_j)$ is the stochastic weighted sum of each adjacency matrix from $\mathcal{P}^*_{bg}$ according to the probability mass vector $w_{i,j}$, defined in Equation \ref{eq:glidr_prob_weights}.
    \begin{equation} \label{eq:soft_adj_mtx}
        \mathbf{M}_{i,j}=\sum_{k=1}^{|\mathcal{P}^*_{bg}|}w_{i,j,k}\gadj_{:,:,k}
    \end{equation}
\end{definition}

\subsubsection{Iterative Message Passing and State Refinement}
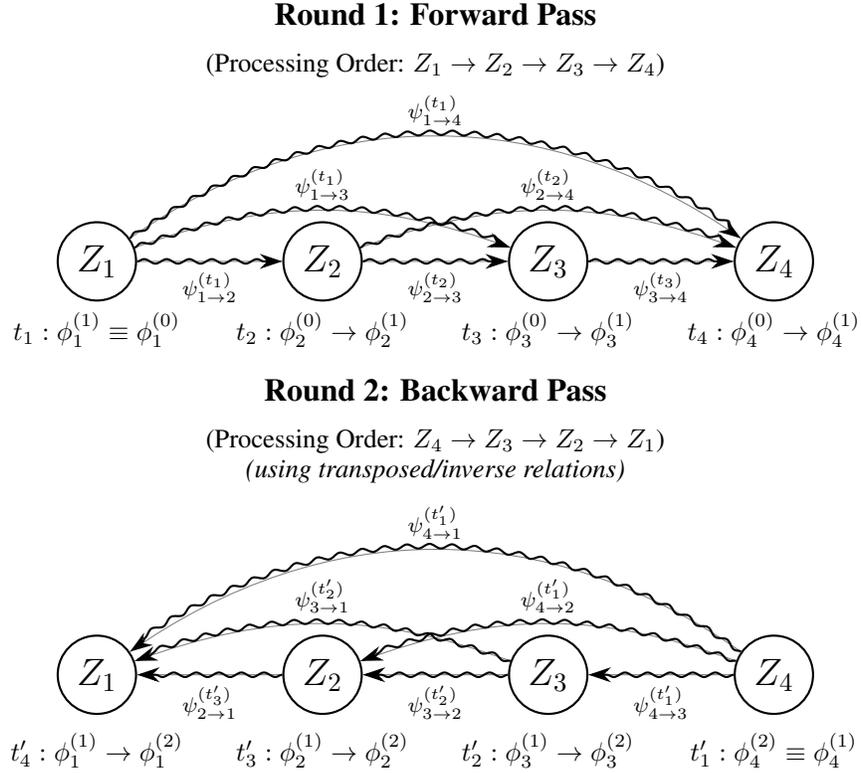
\begin{figure}[htbp]
    \centering
    \begin{tikzpicture}[
        node_style/.style={circle, draw, thick, minimum size=1cm, font=\Large},
        schema_edge_style/.style={draw=gray, thin},
        message_style_r1/.style={-{Stealth[length=3mm, width=2mm]}, draw=black, thick, decorate, decoration={snake, amplitude=0.3mm, segment length=3mm, post length=1mm}},
        message_style_r2/.style={-{Stealth[length=3mm, width=2mm]}, draw=black, thick, decorate, decoration={snake, amplitude=0.3mm, segment length=3mm, post length=1mm}},
        label_style/.style={font=\scriptsize, midway, sloped},
        round_label_style/.style={font=\bfseries\large, anchor=south}
    ]

    \node[round_label_style] (R1_Label) at (4.5, 4.5) {Round 1: Forward Pass};
    \node[below=0.1cm of R1_Label, font=\small] {(Processing Order: $Z_1 \to Z_2 \to Z_3 \to Z_4$)};

    \node[node_style] (R1Z1) at (0,1.5) {$Z_1$};
    \node[node_style] (R1Z2) at (3,1.5) {$Z_2$};
    \node[node_style] (R1Z3) at (6,1.5) {$Z_3$};
    \node[node_style] (R1Z4) at (9,1.5) {$Z_4$};

    \path[schema_edge_style] (R1Z1) edge (R1Z2) (R1Z1) edge [bend left=20] (R1Z3) (R1Z1) edge [bend left=35] (R1Z4)
                             (R1Z2) edge (R1Z3) (R1Z2) edge [bend left=20] (R1Z4)
                             (R1Z3) edge (R1Z4);

    \draw[message_style_r1] (R1Z1) to node[label_style, below] {$\psi_{1\rightarrow2}^{(t_1)}$} (R1Z2);
    \draw[message_style_r1, bend left=20] (R1Z1) to node[label_style, above] {$\psi_{1\rightarrow3}^{(t_1)}$} (R1Z3);
    \draw[message_style_r1, bend left=35] (R1Z1) to node[label_style, above] {$\psi_{1\rightarrow4}^{(t_1)}$} (R1Z4);

    \draw[message_style_r1] (R1Z2) to node[label_style, below] {$\psi_{2\rightarrow3}^{(t_2)}$} (R1Z3);
    \draw[message_style_r1, bend left=20] (R1Z2) to node[label_style, above] {$\psi_{2\rightarrow4}^{(t_2)}$} (R1Z4);

    \draw[message_style_r1] (R1Z3) to node[label_style, below] {$\psi_{3\rightarrow4}^{(t_3)}$} (R1Z4);

    \node[font=\small, below=0.01cm of R1Z1] {$t_1:\phi_1^{(1)}\equiv \phi_1^{(0)} $};
    \node[font=\small, below=0.01cm of R1Z2] {$t_2:\phi_2^{(0)}\rightarrow\phi_2^{(1)}$};
    \node[font=\small, below=0.01cm of R1Z3] {$t_3:\phi_3^{(0)}\rightarrow\phi_3^{(1)}$};
    \node[font=\small, below=0.01cm of R1Z4] {$t_4:\phi_4^{(0)}\rightarrow\phi_4^{(1)}$};

    \node[round_label_style] (R2_Label) at (4.5, -0.5) {Round 2: Backward Pass};
    \node[below=0.1cm of R2_Label, font=\small] {(Processing Order: $Z_4 \to Z_3 \to Z_2 \to Z_1$)};
    \node[below=0.5cm of R2_Label, font=\small\itshape] {(using transposed/inverse relations)};

    \node[node_style] (R2Z1) at (0,-4) {$Z_1$};
    \node[node_style] (R2Z2) at (3,-4) {$Z_2$};
    \node[node_style] (R2Z3) at (6,-4) {$Z_3$};
    \node[node_style] (R2Z4) at (9,-4) {$Z_4$};

    \path[schema_edge_style] (R2Z1) edge (R2Z2) (R2Z1) edge [bend left=20] (R2Z3) (R2Z1) edge [bend left=35] (R2Z4)
                             (R2Z2) edge (R2Z3) (R2Z2) edge [bend left=20] (R2Z4)
                             (R2Z3) edge (R2Z4);
                             
    \draw[message_style_r2] (R2Z4) to node[label_style, below] {$\psi_{4\rightarrow3}^{(t'_1)}$} (R2Z3);
    \draw[message_style_r2, bend right=20] (R2Z4) to node[label_style, above] {$\psi_{4\rightarrow2}^{(t'_1)}$} (R2Z2);
    \draw[message_style_r2, bend right=35] (R2Z4) to node[label_style, above] {$\psi_{4\rightarrow1}^{(t'_1)}$} (R2Z1); 

    \draw[message_style_r2] (R2Z3) to node[label_style, below] {$\psi_{3\rightarrow2}^{(t'_2)}$} (R2Z2);
    \draw[message_style_r2, bend right=20] (R2Z3) to node[label_style, above] {$\psi_{3\rightarrow1}^{(t'_2)}$} (R2Z1);

    \draw[message_style_r2] (R2Z2) to node[label_style, below] {$\psi_{2\rightarrow1}^{(t'_3)}$} (R2Z1); 
    
    \node[font=\small, below=0.1cm of R2Z4] {$t'_1:\phi_4^{(2)}\equiv\phi_4^{(1)}$};
    \node[font=\small, below=0.1cm of R2Z3] {$t'_2:\phi_3^{(1)}\rightarrow\phi_3^{(2)}$};
    \node[font=\small, below=0.1cm of R2Z2] {$t'_3:\phi_2^{(1)}\rightarrow\phi_2^{(2)}$};
    \node[font=\small, below=0.1cm of R2Z1] {$t'_4:\phi_1^{(1)}\rightarrow\phi_1^{(2)}$};

    \end{tikzpicture}
    \caption{Illustration of GLIDR's iterative message passing over two rounds for a 4-node graph-like rule schema ($Z_1, Z_2, Z_3, Z_4$).
    (Top) Round 1: Forward pass, processing variables in order $Z_1 \to Z_4$. Messages $\psi_{i\rightarrow j}^{(t)}$ flow from $Z_i$ to $Z_j$ (where $i<j$).
    (Bottom) Round 2: Backward pass, processing variables in order $Z_4 \to Z_1$. Messages $\psi_{j\rightarrow i}^{(t')}$ flow from $Z_j$ to $Z_i$ (where $i<j$). Messages are computed using using transposed/inverse relations. Each $t_k$ or $t'_k$ indicates a sequential update and message emission step within its respective round.}
    \label{fig:message_passing_rounds}
\end{figure}

GLIDR employs an iterative message-passing procedure to propagate local constraints and refine the soft domains tracked by each variable's state vector $\phi_i$.
Message passing is organized into $R$ rounds that alternate between forward and backward passes through the rule. In each round, there are $N$ time-steps $t$, corresponding to the $N$ variables in the graph-like schema.
After each round, the variables in a rule are implicitly re-indexed in reverse-topological order and the adjacency tensor $B$ is transposed such that $B_{i,j,k}=B_{j,i,k}$. 
Messages in the next round are then effectively sent "backward" in the reverse order of those sent during the previous round. 
This scheduling process ensures that messages are always computed using the most up-to-date information about each variable.
As a result, messages cascade from $(Z_1\rightarrow Z_N)$, $(Z_N\rightarrow Z_1)$, $(Z_1\rightarrow Z_N)$ etc. in alternating "forward" and "backward" passes as depicted in Figure \ref{fig:message_passing_rounds}.
The remainder of this section will assume the indexing scheme of the forward pass.

At each time-step within a round $R$, the value of a state vector $\phi_j^{(R-1)}$ is updated using the set of incoming messages from variables $Z_{i<j}$, and then its new value $\phi_j^{(R)}$ is used to compute new messages to all variables $Z_{k>j}$. Messages are computed and communicated along the "arcs" (predicate slots representing body literals) of the schematic rule structure. Each message $\psi_{i\rightarrow j}\in\mathbb{R}^{|\mathcal{E}_{bg}|}$ encodes a soft belief about which groundings in the domain of $Z_j$ are compatible with the current domain of $Z_i$ according too the predicate modeled by $\mathbf{M}_{i,j}$.

\begin{definition} \label{def:messages}
    A message at time-step $t$ in round $R$, $\psi_{i\rightarrow j}^{(t)}\in\mathbb{R}^{|\mathcal{E}_{bg}|}$, sent from $Z_i\rightarrow Z_j$ is the matrix-vector product of the current state vector for $Z_i$, $\phi_i^{(R)}$, and the soft adjacency matrix $\mathbf{M}_{i,j}$ associated with the $i,j$ slot in the graph-like rule schema. The scalar value $w_{i,j,(-1)}$ holds the probability mass associated with the $P_{true}$ predicate's involvement in the $i,j$ arc.
    \begin{equation} \label{def:messages}
        \psi_{i\rightarrow j}^{(t)}= \mathbf{M}_{i,j}\phi_{i}^{(R)}+\mathbf{1}\cdot w_{i,j,(-1)}
    \end{equation}
\end{definition}
If the soft adjacency matrix $\mathbf{M}_{i,j}$ approximates the adjacency matrix of a specific predicate from $\mathcal{P}_{bg}^*$, then the message $\psi_{i\rightarrow j}$ holds the set of all destination entities in $\mathcal{G}$ of edges originating at the entities encoded in $\phi_{i}$ by the modeled relationship.
Conceptually, this message encodes a soft representation of a set of entities where values crossing some threshold $\epsilon$ are considered present in the set:
\begin{equation}
    \psi_{i\rightarrow j}\approx \{e_j:(e_i,r_{\mathbf{M}_{i,j}},e_j)\in\mathcal{F}_{bg}^*,\forall e_j\text{ s.t. }(\phi_i)_{i}>\epsilon \} 
\end{equation}
The addition of the weight associated with $P_{true}$ ensures that the message becomes $\mathbf{1}$ when all probability mass is placed on $P_{true}$, effectively imposing no constraint on the destination variable's domain.
As in a constraint satisfaction problem, a logic variable in a rule body must have a valid grounding that \emph{simultaneously} satisfies the constraints imposed by all adjacent body literals for a rule to be true.
Determining simultaneous satisfaction involves computing set intersections on the soft set-like state variables $\phi_i$.
We compute approximate set intersection using the element-wise minimum operation.
Before a variable $Z_i$ can emit a message to its neighboring variables $Z_{j>i}$, it must first update its state to reflect any new messages that it received in previous time steps.
\begin{definition} \label{def:state_update}
    A \emph{state update}, $\phi_i^{(R-1)}\rightarrow \phi_i^{(R)}$, for variable $Z_i$ in round $R$, is the element-wise minimum between all incoming messages from variables $Z_{j<i}$ from previous time steps in the round, $\{\psi_{k\rightarrow i}^{(k)}\}_{k=1}^{i-1}$, and the state vector $\phi_{i}^{(R-1)}$.
    \begin{equation} \label{eq:state_update}
        \phi_{i}^{(R)}= \min(\phi_{i}^{(R-1)},\psi_{1\rightarrow i}^{(1)},\ldots,\psi_{(i-1)\rightarrow i}^{(i-1)})
    \end{equation}
\end{definition}
During a forward pass, $Z_1$ has no incoming messages, and Definition \ref{def:state_update} reduces to $\phi_1^{(R)}=\phi_1^{(R-1)}$. The same is true for $Z_N$ during a backward pass.

\subsection{Convergence and Rule Evaluation}
GLIDR's differentiable inference, as described, proceeds through iterative rounds of message passing and variable state refinement. 
This section details how this iterative process terminates, how the final variable states are used to evaluate a rule's entailment for a given query, and the theoretical underpinnings of these outcomes with respect to Constraint Satisfaction Problem (CSP) theory.

\subsubsection{Terminating Inference}
In a "pure" implementation of GLIDR, message passing rounds would continue until one of the following termination conditions was met:
\begin{itemize}
    \item \textbf{Convergence to a Fixed Point:} The iteration is complete if the state vectors $\phi_i$ for all variables $Z_i$ show negligible change between successive rounds. That is, for a round $R$, if $||\phi_i^{(R)}-\phi_i^{(R-1)}||< \epsilon$ for all $i$ and some tolerance $\epsilon$. This indicates that the system has settled into a stable belief about the feasible domains of each variable.
    \item \textbf{Domain Wipeout:} The iteration terminates immediately if the state vector $\phi_i^{(R)}$ for any variable $Z_i$ effectively becomes a zero vector (e.g., $\max(\phi_i^{(R)})\approx 0$). This signifies that no consistent grounding can be found for that variable under the propagated constraints.
\end{itemize}
In practice, we typically set a fixed maximum number of message passing rounds $R_{max}$ to constrain inference complexity.

\subsubsection{Determining Rule Entailment}
After the final round of message passing, GLIDR produces a score, $\hat{y}$, indicating the confidence that the initial grounding query (e.g., $P_h(e_i,e_j)$) is true according to the modeled rule.
\begin{definition} \label{def:rule_entailment}
    A GLIDR model's \emph{confidence score}, $\hat{y}\in(0,1)$, derived from the variable states $\{\phi_i^{(R_{max})}\}_{i=1}^N$ after message passing, represents the model's predicted confidence that the query grounding $P_h(e_h,e_t)$ satisfies the modeled rule for $P_h$.
    \begin{equation} \label{eq:rule_entailment}
        \hat{y} = \min_{1\le i \le N}(\max_{k}(\phi_{i,k}^{(R_{max})})) 
    \end{equation}
\end{definition}
The min-of-maxes form in Definition \ref{def:rule_entailment} ensures that the final predictions produced by GLIDR reflect any domain wipeout that may have occurred during message passing.
If any state variable is close to the zero vector, then the overall score will be close to zero. If all state vectors contain entries close to $1$, then the overall score will be close to $1$.

\subsubsection{Connections to Constraint Satisfaction Theory} \label{sec:theory}
\begin{figure}[htbp]
    \centering
    \begin{tikzpicture}[
        my_node/.style={circle, draw, thick, minimum size=0.8cm, font=\Large, inner sep=1pt},
        my_edge/.style={-{Stealth[length=3mm, width=2mm]}, draw=black, thick},
        z_label_style/.style={below=0.2cm, font=\Large, text=black}
    ]
    \node [my_node] (n0) at (-2, 0) {$e_0$};
    \node [my_node] (n1) at (-2, 1) {$e_1$};
    \node [my_node] (n2) at (2, 0) {$e_2$};
    \node [my_node] (n3) at (2, 1) {$e_3$};
    \node [my_node] (n4) at (0, -1) {$e_4$};
    \node [my_node] (n5) at (0, -2) {$e_5$};
    \node [my_node] (n6) at (4, 0.5) {$e_j$};
    \node [my_node] (n7) at (-4, 0.5) {$e_i$};

    \draw [my_edge] (n7) to (n1);
    \draw [my_edge] (n1) to (n3);
    \draw [my_edge] (n3) to (n5);
    \draw [my_edge] (n5) to (n0);
    \draw [my_edge] (n0) to (n2);
    \draw [my_edge] (n2) to (n4);
    \draw [my_edge] (n4) to (n1);
    
    \draw [my_edge] (n3) to (n6);
    \draw [my_edge] (n2) to (n6); 
    \draw [my_edge] (n7) to (n0);

    \node [z_label_style] at (n7.south) {$Z_1$};
    \node [z_label_style] at (n0.south) {$Z_2$};
    \node [z_label_style] at (n2.south) {$Z_4$};
    \node [z_label_style] at (n5.south) {$Z_3$};
    \node [z_label_style] at (n6.south) {$Z_5$};

    \end{tikzpicture}
    \caption{A simple counterexample illustrating variable domains that are \emph{locally consistent} but not \emph{globally satisfiable}. Each stack of entities is the domain $D(Z_i)$ for a logic variable in the rule $P_h(Z_1,Z_5)\leftarrow P_{1,2}(Z_1,Z_2)\wedge P_{2,4}(Z_2,Z_4) \wedge P_{4,3}(Z_4,Z_3) \wedge P_{3,2}(Z_3,Z_2) \wedge P_{4,5}(Z_4,Z_5)$. The arrows represent background facts satisfying the constraints imposed by each body literal. The domains are an entirely arc consistent fixed point, but the modeled query is not true.}
    \label{fig:counterexample}
\end{figure}
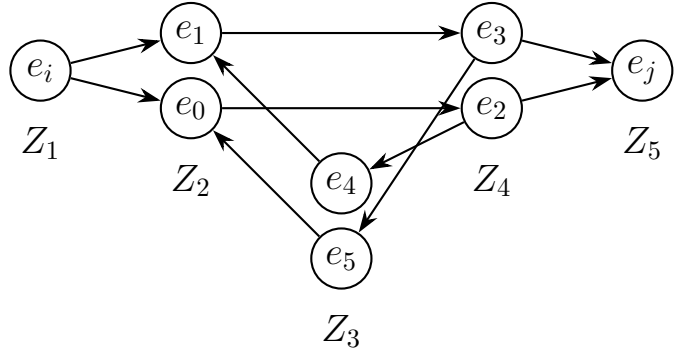

The message passing inference algorithm that GLIDR uses to determine rule satisfiability can be interpreted as a form of iterated local consistency enforcement. 
Iterative local consistency is the same core inference mechanism used by backtrack-free algorithms like AC-3 \cite{mackworth_ac3} to solve constraint satisfaction problems.
Local consistency algorithms are characterized by applying constraints locally between variables and iterating until convergence.
The constraints imposed on variable groundings by body literals in GLIDR are analogous to arc consistency enforcement \cite{mackworth_ac3}.
CSP theory offers several theoretical results that can applied to GLIDR's inference in the hard setting. In the soft setting, GLIDR's inference is already fully approximate.
\begin{itemize}
    \item \textbf{Domain Wipeout Implies Falsity:} The occurrence of a domain-wipeout during local consistency propagation guarantees unsatisfiability of the constraint network \cite{russel_ai_modern_approach}. The converse is not generally true.
    \item \textbf{Fixed-Point Proves Satisfiability for Tree-Like Networks:} If a constraint network is structured like a tree, then arrival at a fixed point (without domain collapse) during arc consistency propagation is sufficient to prove satisfiability \cite{Kumar_1992,freuder_suff_cond}. This implies that GLIDR's inference can be exact for cycle free rules such as chain-like and tree-like rules.
    \item \textbf{Fixed-Point in Loopy Network Does Not Prove Satisfiability:} In a loopy network, it is possible to arrive at a fixed point solution that is \emph{locally consistent}, but still globally unsatisfiable. Proving satisfiability in such a case requires backtracking, and GLIDR's inference will be incorrect. Figure \ref{fig:counterexample} illustrates such an example.
\end{itemize}

\subsection{Optimization}

GLIDR models are essentially binary classifiers that accept a background graph and a pair of entity indices to make predictions. 
This means that, unlike Neural-LP and DRUM \cite{yang2017differentiable,sadeghian2019drum}, GLIDR requires negative examples during training.
We sample positive examples of entities that share the target relationship to form mini-batches. 
We employ the closed-world hypothesis during negative example selection, sampling entity pairings that, to the best knowledge in the training data, do not share the target relationship.
The sampling rates of positive and negative examples are weighted to ensure an average 1:1 ratio of positives and negatives in a mini-batch during training.
GLIDR is optimized to maximize its confidence score when applied to a positive example and minimize its confidence score when applied to a negative example. 
We adopt the pairwise logistic loss \cite{burges2005ranking} when training \method\ for ranking problems because it promotes high relative score difference between positive and negative examples in a batch, even when a modeled rule is unsatisfiable and producing low confidence scores. 
\begin{definition}
The \emph{pairwise logistic loss} \cite{burges2005ranking} for a batch of predictions $\hat{\mathbf{y}}$ and score labels $\mathbf{y}$ is defined by Equation \ref{eq:pairwise_log_loss}. $\mathbb{I}[y_i>y_j]$ is the indicator function and takes a value of 1 when $\mathbf{y_i}>\mathbf{y_j}$ and 0 otherwise.
\begin{equation} \label{eq:pairwise_log_loss}
    l(\hat{\mathbf{y}},\mathbf{y})=\sum_i\sum_j\mathbb{I}[\mathbf{y_i}>\mathbf{y_j}]\log(1+\exp(-(\hat{\mathbf{y}}_i-\hat{\mathbf{y}}_j)))
\end{equation}
\end{definition}
 The pairwise logistic loss is constructed for problems with continuous labels, but we can apply it to problems like ours with discrete labels $y\in\{0,1\}$.

We typically learn multiple rule definitions for each target relationship when training \method\ on a dataset. A set of weights is randomly initialized for each rule definition at the beginning of training, and each rule produces predictions during training. For a given training batch, the pairwise logistic loss is applied to each rule's predictions and the resulting loss values are averaged together. 
We use the AdamW optimizer \cite{adamw2017} to train GLIDR, and we find that the method converges best with high learning rates in the vicinity of $0.1$. We also find that GLIDR benefits from a high weight decay, also in the vicinity of $0.1$.

\subsection{Rule Extraction}
The approximate rules learned by GLIDR in the soft setting must be \emph{extracted} before they can be used by a symbolic solver. The process of rule-extraction is crucial to the interpretability of differentiable rule learning methods, and many differentiable rule learning methods approximate rules that are difficult to explicitly extract \cite{faithfulre}.

Chain-like differentiable ILP techniques like Neural-LP and DRUM \cite{yang2017differentiable,sadeghian2019drum} derive rule satisfiability by \emph{counting} paths on the background graph and mix different sequential reasoning pathways to produce a confidence score. This can make the extraction of rules that correctly reflect the behavior of the soft models difficult \cite{faithfulre}.
GLIDR mitigates some of these issues by maintaining separate and explicit pathways for interaction between variables.
GLIDR's inference process is also more closely related to traditional logical inference, seeking a mutually consistent state across the soft domains that it tracks.
The constraint that these soft domains are bounded on $[0,1]$ prevents GLIDR from "counting" the frequency of entities in a domain (at least in the hard setting).

A key consideration in extracting symbolic rules from GLIDR arises when the learned probability distribution $w_{i,j}$ for predicate slot $(i,j)$ does not sharply concentrate on a single predicate. This indicates model uncertainty, or that multiple predicate types contributed to the learned behavior for that slot. GLIDR's construction of the soft adjacency matrix for a slot, Equation \ref{eq:soft_adj_mtx}, is inherently additive. This means that when multiple predicates $P_k$ have substantial weights $w_{i,j,k}$, the inference step involving $\mathbf{M}_{i,j}$ effectively aggregates the outcomes as if \emph{each} of these highly weighted predicates is "active" simultaneously. Consequentially, the behavior of such a slot can be interpreted as a soft logical disjunction of the highly weighted predicates $\bigvee_kP_k$. While the primary goal of GLIDR is to learn Horn clauses without disjunction among body literals, sometimes extracted rules can be more faithfully represented by introducing disjunctions.

One heuristic that we explore for rule extraction is a "top $p$" sampling of each predicate distribution. 
All predicates with weights in the top $p$ cumulative probability mass are extracted and added to a disjunctive body term.
If $P_{true}$ falls within the top $p$ probability mass, then $P_{true}$ is sampled for that slot instead of any other predicates.
This heuristic has the desirable property that a highly converged soft model will often produce the same hard rule with top $p$ as with an argmax strategy (where the highest scoring predicate is assigned to each slot).
This means that in highly converged models, the resulting rule is purely conjunctive.
When a model shows lower convergence, a limited number of disjunctive body literals are introduced based on the parameter $p$.
If $p$ is low, then a limited number of disjunctive body literals are added to a rule. If $p$ is high, then a large number of disjunctive body literals can be added to a rule.
The inclusion of many disjunctive body literals can hurt rule interpretability by introducing many possible truth conditions.
In testing different rule extraction heuristics, we observed that no single heuristic worked best across all rules.
This indicates that there is likely significant performance to be gained by applying a search-based approach to rule extraction.

During evaluation, we performed a top $p=0.25$ rule extraction on GLIDR models trained on multiple datasets. We observed that the disjunctions introduced were sometimes informative and reflected disjunctive patterns in the background data. For example, this rule for \emph{niece} was extracted from the family dataset:
\begin{equation*}
    \text{niece}(X,Y)\leftarrow \text{niece}(X,Z_1)\wedge(\text{aunt}(Y,X)\vee \text{uncle}(Y,X))\wedge \text{brother}(Y,Z_1).
\end{equation*}
While this rule isn't perfect, mishandling the scenario where $Y$ is an aunt of $X$, it does give us a glimpse into the internal tension within the model. The "aunt or uncle" disjunction will explain every instance of niece, but it will also explain every instance of nephew. To control the gender of $X$, the model has included the niece and brother terms, however they constrain $Y$ to be an uncle, hurting the generality of the rule. The model seems to be splitting the difference between a rule that fully commits to classifying an uncle-niece relationship, and a rule that classifies an aunt-niece relationship. In theory, a dangling niece$(X,Z_2)$ term with no corresponding brother would be sufficient to make this an acceptable hard rule, however the split probability mass between the aunt and uncle predicates limits the maximum score of such a rule. This may indicate that more explicit handling of disjunctive terms could benefit future methods.

\subsection{Complexity}
When run until convergence, GLIDR's worst-case inference compute complexity is $\mathcal{O}(N^3D^3)$. Where $N$ is the number of schematic variables, and $D=|\mathcal{E}_{bg}|$ is the maximum domain size (background entity count). Computing each message involves a matrix-vector product which is $\mathcal{O}(D^2)$. There are $\mathcal{O}(N^2)$ messages passed in each round of message passing. If any round of message passing concludes without introducing a change in a variable domain, then a fixed-point has been reached and inference is converged. There are $\mathcal{O}(ND)$ total possible entities represented in all variable domains. In the worst-case, where only a single entry in a single domain is eliminated per round of message passing, it would take $\mathcal{O}(ND)$ rounds of message passing to reach a domain wipeout. This yields the overall worst-case complexity $\mathcal{O}(N^3D^3)$. The high-degree of sparsity in graph datasets (e.g. $\mathbf{B}$ is highly sparse) means that most messages are also highly sparse, leading to rapid domain convergence. The use of sparse linear algebra primitives also constrains the typical complexity of the matrix-vector product required for each message. GLIDR inference is typically only performed for a fixed $R_{max}$ rounds, so the typical inference compute complexity \emph{per-query} $P_h(e_i,e_j)$ is $\mathcal{O}(N^2D^2)$. In the case of open queries like $P_h(e_i,Y)$, $D$ inferences must be made for the $D$ possible tail entities. This can be quite expensive as $|\mathcal{E}_{bg}|$ grows large.
\section{Experiments}

To assess the graph learning capabilities of \method, we evaluate its performance on several knowledge graph completion benchmarks and subject it to mislabeling noise. Following the approach used by Neural-LP \cite{yang2017differentiable}, each dataset is partitioned into four splits: \emph{train}, \emph{validation}, \emph{test}, and \emph{facts}. During training, the \emph{facts} split is used to construct the background knowledge graph, $\mathcal{G}_{bg}$, for rule inference. During validation and testing, both the \emph{facts} and \emph{train} splits are combined to form the background knowledge graph used for reasoning. The \emph{facts} and \emph{train} splits were formed by partitioning the \emph{train} split commonly used by embedding methods \cite{yang2017differentiable}. \method\ fits rule definitions during the training stage and evaluates them during the validation and testing stages; therefore, we require that the sets of predicates in each split overlap. The rule definitions fit by \method\ only involve lifted predicates and logic variables, so in contrast with many embedding methods, we do not require that the entities in any split overlap. In this sense, the rules learned by GLIDR are entirely \emph{inductive}, and can be applied to any dataset with the appropriate predicates.

We implement \method\ in JAX \cite{jax2018github} using the experimental sparse module to perform sparse tensor operations. Our implementation also makes use of sharding and parallelism primitives provided by JAX to enable multi-gpu training with many rule bodies using model parallelism. We use the implementations of the pairwise logistic loss \cite{burges2005ranking} and ranking utilities provided by Rax \cite{jagerman2022rax}. Our experiments use the Optax \cite{deepmind2020jax} implementation of the AdamW optimizer \cite{adamw2017}.

\subsection{Knowledge Graph Completion}

We evaluate \method\ on the knowledge graph completion problem described by \cite{bordes2013translating}. In this problem, a relational query of the form \texttt{relation(head,?)} is posed, and a system must find tail entities that complete the query by ranking all entities in the background graph by the likelihood that the fact \texttt{relation(head,tail)} exists in the knowledge base.
To evaluate knowledge graph reasoning systems in this manner, a held-out set of facts from a knowledge base is used to generate queries, and systems must rank the entities that complete these held-out facts as highly as possible. 

Bordes et al. establish several conventions for evaluating methods on this task. 
Queries are constructed to retrieve both head entities \texttt{relation(?,tail)} and tail entities \texttt{relation(head,?)} for each fact in the evaluation set. 
Retrieval metrics such as hits@$k$, which count the fraction of target entities that are ranked within the top $k$ results for each query are used to report performance. 
We adopt the filtered setting described by Bordes et al. in which true positive completions do not affect the rank of entities ranked below them. 
This setting modifies the interpretation of hits@$k$ to be the proportion of target entities that are ranked below fewer than $k$ false positives.
A further issue with ranking metrics arises when rankings have the potential for ties as is common with methods that assign numerical scores.
Optimistic or pessimistic tie-breaking can have significant impacts on ranking evaluation as noted by \cite{sun2020evaluation}, so we follow the recommendation of applying random tie-breaking.

For each dataset, we learn a collection of 8 rule definitions for each predicate. We initialize GLIDR with $N=4$ schematic variables, implying that the maximum depth of any learned rule is 3. This length restriction is consistent with the other methods we evaluate against. Following other rule learning methods \cite{yang2017differentiable, qu2020rnnlogic, cheng2023neural} we allow \method\ to learn recursive rule definitions during benchmarking.
To avoid the possibility of recursive rules exploiting the facts that they are meant to predict, we ensure that we only train on positives from the \emph{train} split, while negatives can be sampled from either the \emph{train} or \emph{facts} splits. This ensures that a positive training example never exists in the background data.

To facilitate better rankings, we create a weight for each rule definition based on that rule's performance on the validation set. Each definition is assigned a weight equal to the ratio of the minimum average validation loss across definitions to its own average validation loss $w_i=\frac{\bar{l}_\text{min}}{\bar{l}_i}$. The best performing definitions get a weight of $1.0$ and worse performing definitions get down-weighted. The final ranking score generated by \method\ is the weighted sum of the confidence scores produced by each rule definition.

\subsection{Datasets}
We train and evaluate \method\ on the knowledge graph completion task for the Family \cite{yang2017differentiable}, Alyawarra Kinships \cite{denham1973, denham2016}, UMLS \cite{mccray2003umls, bodenreider2004umls}, and Freebase15k-237 \cite{toutanova2015observed} datasets. For each of these datasets, we use the splits provided by Neural-LP \cite{yang2017differentiable}. Family is a noisy dataset with 3007 entities related by 12 western kinship terms following the small Kinship dataset \cite{hinton1990kinship}. The Unified Medical Language System (UMLS) is a dataset of 135 biomedical entities and 46 relational terms. The Alyawarra Kinships dataset (henceforth referred to as Kinships), is a dataset derived from ethanographic data collected by Denham in 1973. This dataset was constructed by first photographing 104 Alyawarra-speaking Aboriginal people of the central Australian outback. Participants were then shown pictures of the other 103 persons photographed and asked to name a kinship relation for each relative to themself. The resulting graph is fully connected and contains 25 different Alyawarra kinship terms. Freebase15k-237 is a subset of the Freebase knowledge graph containing 14541 entities and 237 relation types that primarily describe sports, media, and geographical concepts.

\subsection{Existing Methods}

We compare \method\ to a selection of rule learning and embedding methods. Among embedding methods, we choose to compare against ConvE \cite{dettmers2018conve}, an algorithm based on graph convolutions, and RotatE \cite{sun2019rotate} an algorithm that makes use of complex vector space embedding transformations and an improved adversarial negative sampling technique. We also compare against RNNLogic \cite{qu2020rnnlogic}, a hybrid rule-based and embedding method that learns chain-like rules. The first reported setting for RNNLogic makes use of both logical rules and learned entity embeddings. The second setting, denoted by \texttt{w/o emb.} is only rule-based and does not make use of embeddings. In addition to RNNLogic, we compare against NeuralLP \cite{yang2017differentiable}, a differentiable chain-like rule learning method. DRUM \cite{sadeghian2019drum} improves on NeuralLP by modifying the mechanism for variable-length rule learning. We also include NCRL \cite{cheng2023neural}, another rule based method that learns compositional chain-like logic rules and scales well to large datasets.

\begin{table*}[]
\begin{center}
\begin{tabular}{c p{.5cm} p{.8cm} p{1cm} p{.5cm} p{.8cm} p{1cm} p{.5cm} p{.8cm} p{1cm} p{.5cm} p{.8cm} p{1cm}}
\hline
Model & \multicolumn{3}{c}{Family} & \multicolumn{3}{c}{Kinships} & \multicolumn{3}{c}{UMLS} & \multicolumn{3}{c}{FB15k-237} \\
 & MRR & Hits@1 & Hits@10 & MRR & Hits@1 & Hits@10 & MRR &  Hits@1 & Hits@10& MRR &  Hits@1 & Hits@10 \\ \hline
ConvE & - & - & - & \textbf{0.83} & \textbf{73.0} & \textbf{98.0} & \textbf{0.94} & \textbf{92.0} & \textbf{99.0}& 0.32 & 24.0 & 49.0 \\ 
RotatE & \textbf{0.86}$^\dagger$ & \textbf{78.7}$^\dagger$ & \textbf{93.3}$^\dagger$ & 0.65 & 50.4 & 93.2 & 0.74 & 63.6 & 93.9 & 0.34 & 24.1 & \textbf{53.3} \\
RNNLogic & - & - & - & 0.72 & 59.8 & 94.9 & 0.84 & 77.2 & 96.5 & \textbf{0.34} & \textbf{25.2} & 53.0 \\ \hdashline
w/o emb. & 0.86$^\dagger$ & 79.2$^\dagger$ & 95.7$^\dagger$ & 0.64 & 49.5 & 92.4 & 0.75 & 63.0 & 92.4 & 0.29 & 20.8 & 44.5 \\
Neural-LP & 0.88$^\dagger$ & 80.1$^\dagger$ & 98.5$^\dagger$ & 0.30$^\ddagger$ & 16.7$^\ddagger$ & 90.1 & 0.48$^\ddagger$ & 33.2$^\ddagger$ & 93.2 & 0.24 & 17.3$^\ddagger$ & 36.2\\ 
DRUM (L=3) & \textbf{0.95} & 91.0 & 99.0 & 0.61 & 46.0 & 91.0 & 0.80 & 66.0 & \textbf{97.0} & \textbf{0.34} & \textbf{25.5} & \textbf{51.6} \\ 
NCRL & 0.91 & 85.2 & \textbf{99.3} & 0.64 & 49.0 & 92.9 & 0.78 & 65.9 & 95.1 & 0.30 & 20.9 & 47.3 \\ 
Soft ($\mu$) & 0.90 & \textbf{93.2} & 95.6 & \textbf{0.72} & \textbf{73.5} & \textbf{93.1} & \textbf{0.81} & \textbf{87.8} & 95.2 & 0.20 & 18.6 & 35.6\\ 
Soft ($\sigma$) & 0.01 & 0.67 & 0.72 & 0.01 & 1.09 & 0.92 & 0.01 & 0.63 & 0.37 & - & - & -\\ 
Top $p=0.25$ ($\mu$) & 0.66 & 68.4 & 72.4 & 0.61 & 61.7 & 78.3 & 0.67 & 71.2 & 87.0 & - & - & - \\ 
Top $p=0.25$ ($\sigma$) & 0.04 & 4.75 & 5.42 & 0.01 & 1.53 & 1.77 & 0.02 & 2.82 & 2.20 & - & - & -\\ \hline 
\end{tabular}
\label{tab:metrics}
\caption{\textbf{Performance of embedding and rule-based methods on several knowledge graph completion benchmarks.} Embedding methods are reported above the first dashed line and rule-based methods below. The best result for each metric is reported independently in \textbf{bold} for embedding methods and rule-based methods. We report mean ($\mu$) and standard deviation ($\sigma$) scores for GLIDR in both the hard ("Top $p$") and soft ("Soft") settings across 10 runs for Family, Kinships, and UMLS. All scores for other differentiable rule learning systems are reported for the soft setting. Hits@$k$ is reported in \%. $[\dagger]$ indicates a result taken from the NCRL paper \cite{cheng2023neural}. $[\ddagger]$ indicates a result from the RNNLogic paper \cite{qu2020rnnlogic}. All other results are from the original papers.}
\end{center}
\end{table*}
\subsection{Results}
We compare mean reciprocal rank (MRR), Hits@1, and Hits@10 for \method\ and the other selected algorithms in Table \ref{tab:metrics}. For Family, Kinships, and UMLS, we train and evaluate \method\ 10 times with different random seeds and report the mean and standard deviation for each metric. Generating tail rankings for all test queries on FB15k-237 is very expensive given GLIDR's ground-and-check inference algorithm, and we only report the result from a single run. Compute resources and hyperparameters for each experiment are reported in the appendix \ref{sec:hyperparams}.

We observe that \method\ performs very competitively with other rule learning methods. On Kinships, UMLS, and Family, \method\ has the top performing Hits@1 among rule learning methods, and its scores are often competitive with embedding methods. GLIDR does not significantly improve on Neural-LP's performance on FB15k-237. We hypothesize that the structure of the dataset may not benefit significantly from the added expressivity that GLIDR's rules offer. FB15k-237 is also significantly larger than the other datasets studied, and it is possible that this difference necessitates special treatment with modified training hyperparameters.

Generally, we observe that our method performs better in Hits@1 than MRR or Hits@10. We believe that the highly-constrained nature of GLIDR's rules produces sharper drops in confidence scores when entity pairings do not perfectly satisfy a rule. This allows learned rules to be highly selective in their decision criteria at the cost of covering fewer cases.

It is important to note, as previous authors have \cite{sadeghian2019drum}, that rule based methods cannot be fairly compared with embedding methods, since embedding methods can store per-entity information at training time and utilize it at test time. Previous authors have shown that the performance of embedding methods suffers significantly in the purely inductive setting where no entities are shared between training and test time, while rule-based methods can retain much of their performance \cite{yang2017differentiable,sadeghian2019drum}.

We tested hard rules extracted from GLIDR's soft weights using the top $p=0.25$ rule extraction heuristic on Kinships, UMLS, and Family. Inference was performed using GLIDR in the hard setting, with binary probability weights $w_{i,j}$. We used the same weighted averaging based on validation loss as in the soft setting. We found that rules extracted from GLIDR retained a large proportion of the performance of the original soft models. In Kinships and UMLS, the soft rules still outperformed the next best rule learning method in Hits@1.

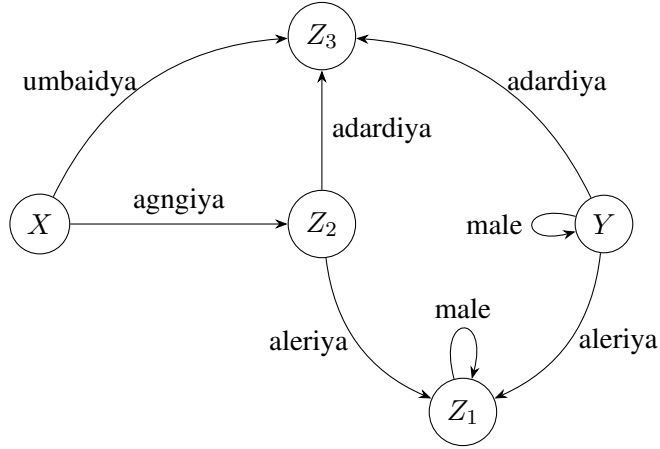
\begin{figure}
    \centering
    \begin{tikzpicture}
      \node (n1) at (0,0) [circle, draw, minimum size=.5cm] {$X$};
      \node (n2) at (5.625,-2.5) [circle, draw, minimum size=.5cm] {$Z_1$};
      \node (n3) at (3.75,0) [circle, draw, minimum size=.5cm] {$Z_2$};
      \node (n4) at (3.75,2.5) [circle, draw, minimum size=.5cm] {$Z_3$};
      \node (n5) at (7.5,0) [circle, draw, minimum size=.5cm] {$Y$};
    
      \draw[-{Stealth}, bend left=0] (n1) to node[above, pos=0.5] {agngiya} (n3);
      \draw[-{Stealth}, bend left=30] (n1) to node[left, pos=0.5] {umbaidya} (n4);
      \draw[-{Stealth}, bend right=30] (n3) to node[left, pos=0.5] {aleriya} (n2);
    
      \draw[-{Stealth}, bend left=0] (n3) to node[right, pos=0.5] {adardiya} (n4);
      \draw[-{Stealth}, bend right=30] (n5) to node[right, pos=0.5] {adardiya} (n4);
    
      \draw[-{Stealth}, bend left=30] (n5) to node[right, pos=0.5] {aleriya} (n2);
      \draw[-{Stealth}, out=165, in=195, looseness=10] (n5) to node[left, pos=0.5] {male} (n5);
      \draw[-{Stealth}, out=105, in=75, looseness=10] (n2) to node[above, pos=0.5] {male} (n2);
    \end{tikzpicture}
    \caption{Illustration of the learned rule graph for a recursive rule discovered for Agngiya in the original Kinships dataset.}
    \label{fig:agngiya-graph}
\end{figure}

In keeping with the expressive nature of its rule parameterization, we observe that \method\ learns collections of both chain-like and graph-like rules for each dataset. One example of a graph-like rule extracted from \method\ during an experiment with the original Kinships data is visualized in Figure \ref{fig:agngiya-graph}. This recursive rule for Agngiya was discovered from a schematic rule with $N=6$. We found that this extracted rule achieved an F1 score of 0.803 when used to predict the existence of Agngiya relationships in the graph. This was a transductive setting, where all background data was available at training and test time.

\subsection{Performance Subject to Noise}

\begin{figure}
    \centering
    \includegraphics[width=.7\textwidth]{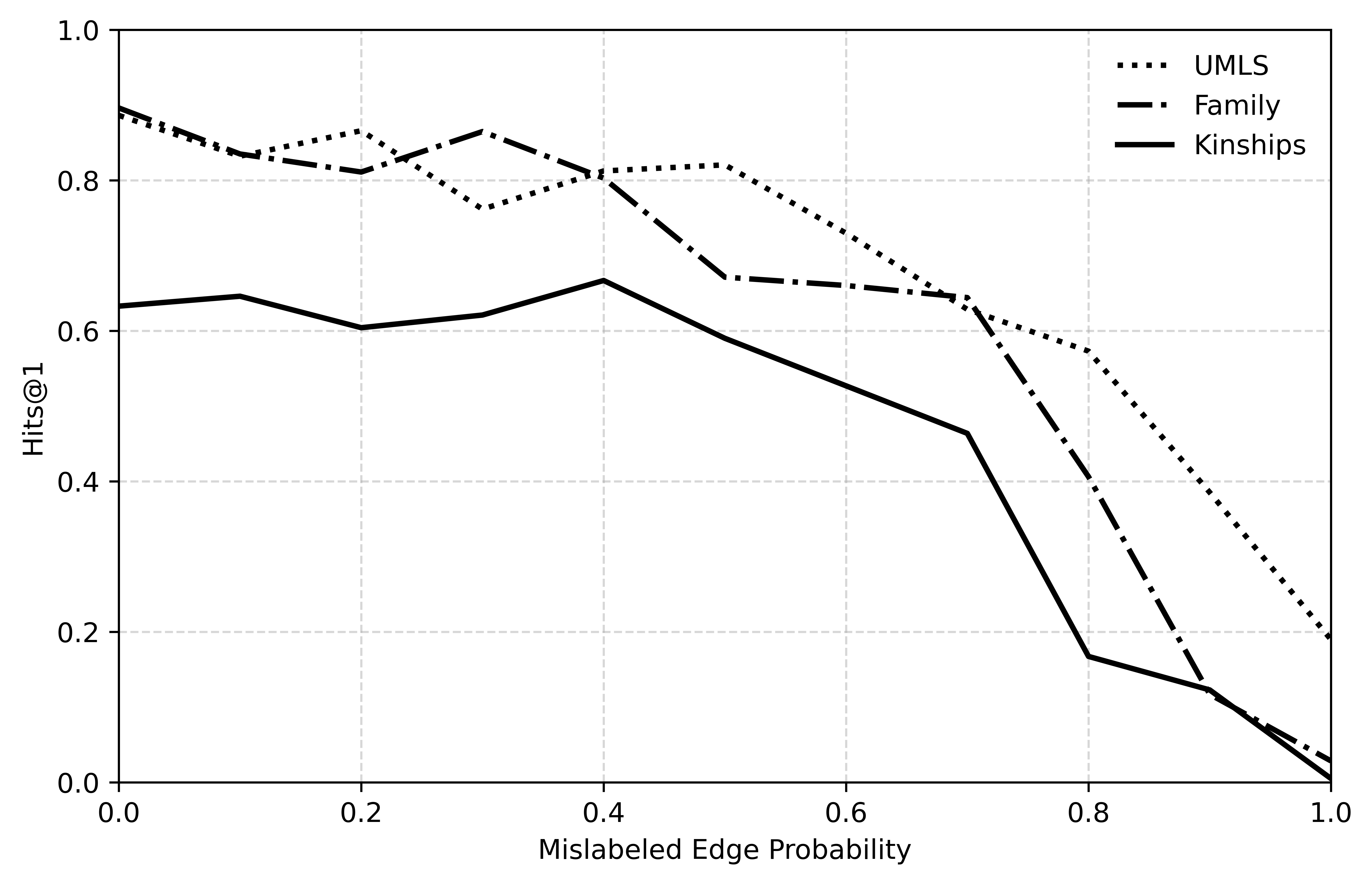}
    \caption{Hits@1 on the Kinships, Family, and UMLS datasets as the probability of edge mislabeling $p$ increases.}
    \label{fig:mislabeling}
\end{figure}

To investigate the performance impact that noisy data has on \method, we conduct an experiment where we purposefully introduce mislabeled edges into the training dataset. With some probability $p$, we change the predicate type of each edge in the \emph{train} and \emph{facts} splits to a different predicate type during training. Although some edges may still match a fact from the unperturbed data after mislabeling, the widespread random relabeling of edges should introduce significant noise into the training graph. We evaluate \method\ on Kinships, Family, and UMLS while varying the probability of mislabeling from $p=0$ to $p=1$. We report Hits@1 for each of these trials in Figure \ref{fig:mislabeling}. We find that our method shows significant robustness to mislabeling across all three datasets. In each case, \method\ only incurs minor performance degradation at mislabeling rates below 50\%. We also observe a steady decline in performance as mislabeling increases to 100\% rather than a sudden drop, indicating that the method is consistently robust to noise even at high rates.

\subsection{Choosing the Number of Schematic Variables}

\begin{figure}
    \centering
    \includegraphics[width=.7\textwidth]{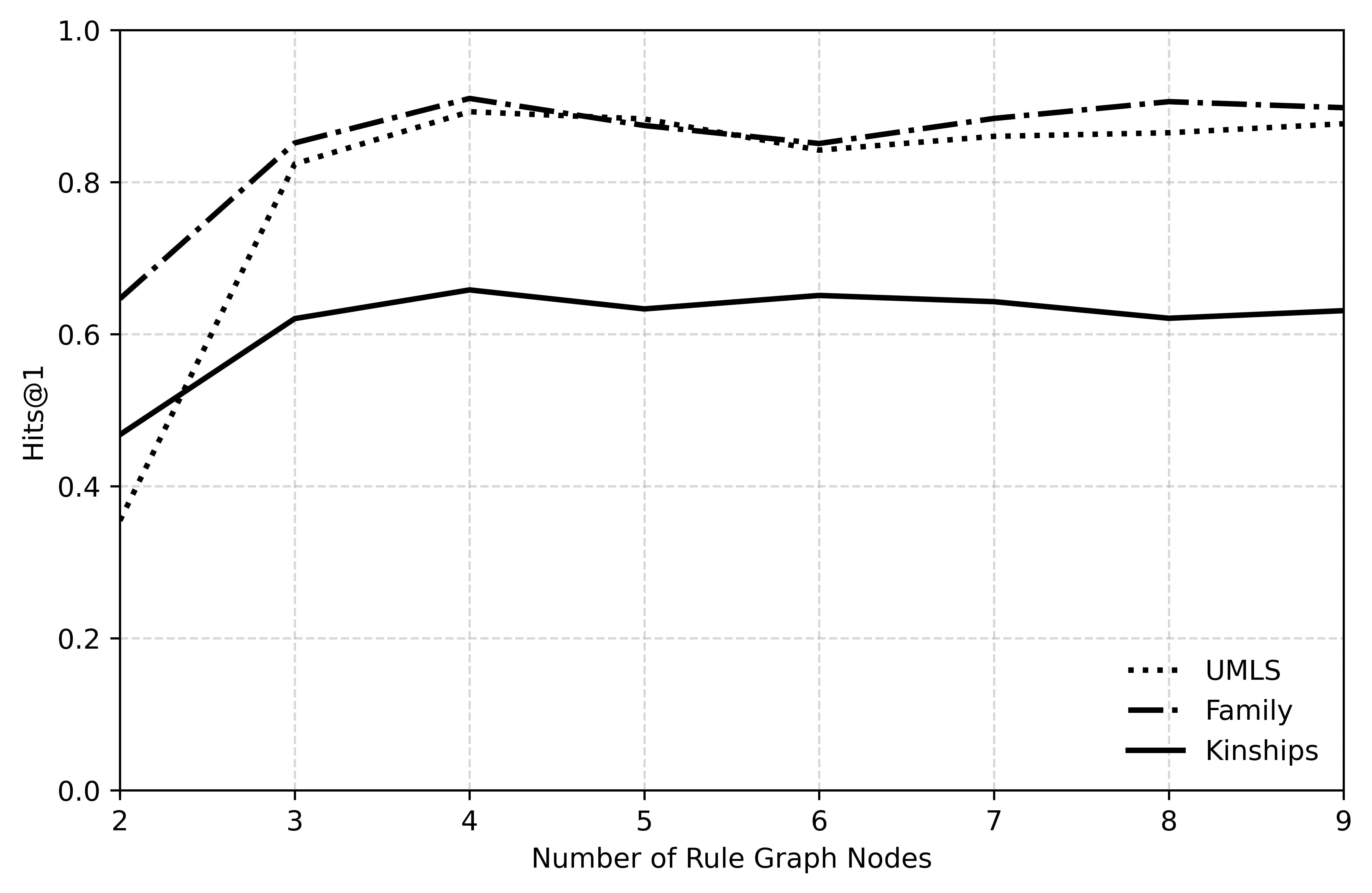}
    \caption{Hits@1 on the Kinships, Family, and UMLS datasets as the number of nodes in the schematic rule, $N$, grows.}
    \label{fig:rule_length}
\end{figure}

In theory, \method\ parameterized with $N$ schematic variables should be able to learn any rule that can be expressed with $n'<N$ variables. To investigate whether this happens in practice, we again trained \method\ on Kinships, Family, and UMLS. For each dataset, we varied the number of schematic variables from $N=2$ to $N=9$ and recorded the results. Figure \ref{fig:rule_length} plots the Hits@1 on each dataset as the number of variables was changed. We observe that there is a performance benefit to increasing the rule graph size up to a certain point, after which the performance plateaus. This can be explained by the existence rules of a certain length which are sufficient to predict the relationships in a dataset. As the rule graph grows past that size, the system continues to learn the optimal rules as smaller and smaller subsets of the larger schematic rule. This is a very good property, as it implies that a user can set the rule graph as large as their computer and patience allows without fear of hurting model performance.

\subsection{End-to-End Co-Training with Deep Learning Models}
To demonstrate the feasibility of using \method\ in a larger deep learning system, we co-train \method\ with an image encoder to classify MNIST \cite{lecun2010mnist} digits. 
We choose to use MNIST as a test-case to simplify the analysis of predicates learned during the co-training process. 
In this experiment, images are divided into a 7x7 grid of 4x4 pixel image patches. Each image patch is independently vectorized and mapped into a latent representation using a four layer MLP. 
We model each image as a 49 node multi-graph, where each node corresponds to a patch in the original image. We choose a number of binary and unary predicates to generate for inclusion in this background graph. 
Each binary relationship and unary property is generated using the embedded representations of the image patches. To create unary facts, a linear projection and sigmoid activation function are applied to each image patch embedding. 
To create relational facts, each image patch embedding is projected into query and key embeddings and these are combined with an attention mechanism \cite{vaswani2017attention}. 
Rather than using a softargmax in the attention mechanism, we use a sigmoid function to produce a soft adjacency matrix for each predicate type. We manually zero the diagonals of these matrices and clip any values below 0.5 to zero. Clipping small activations to zero promotes sparsity in the adjacency matrices, and we find that this is a key modification required to achieve convergence during training. 
With a trainable mapping from images to graph adjacency arrays and node property arrays, we can model the digit classification problem as a graph classification problem with GLIDR. 
We initialize GLIDR with 10 body rules, one for each digit type. We take the scores produced by each body rule and apply a softargmax, so that the "most satisfied" rule is the classification decision made by the ensemble. 
We then train the GLIDR model as usual, taking batches of images, mapping them into graphs, and then passing those graphs into GLIDR. 

We observe that the high learning rates favored by GLIDR are unsuited for training the image encoder. To address this, we use two AdamW optimizers with different hyperparameters. Gradients are backpropagated end-to-end through the system, and a high learning rate AdamW is used to update GLIDR's parameters while a low learning rate AdamW is used to update the parameters of the image encoder. 
With this setup, we observe rapid convergence of the combined models. With an $N=4$ GLIDR backbone, 8 binary predicates, 8 unary predicates, and a 4-layer MLP, we achieve an F1 Score of 0.79 on the MNIST test set. 
We find that the performance of the system improves as the number of available predicates increases, with 32 binary predicates and 32 unary predicates achieving an F1 score of 0.88. 

Notably, no information about the spatial layout of the image patches is retained in this system, and we expect some inherent limitation on the performance of this system due to the constrained nature of viewing images as a bag of patches. 
Despite this, we observe that many of the learned predicates attempt to recover elements of the original image structure. For example, one learned predicate (heavily featured in the rule for 1) consistently maps left edges of vertical lines to similar looking right edges of vertical lines. In doing so, this predicate measures the prevalence of vertical edges in the image. Indeed, we observe that the learned definition of "1" makes exclusive use of predicate 6 for its binary relationships. 
Similarly, we observe that another predicate maps tops and bottoms of horizontal edges together to recover information about the existence of horizontal lines. 

For the smaller experiment with 8 predicates, we observe a high degree of predicate re-use between the rules for different digits. As the domain grew to 32 predicates, few were re-used across rules. This indicates that successful discovery of general purpose predicates might require a learning schedule or another mechanism to promote predicate reuse.

The rules learned by this system are not directly interpretable due to their use of learned neural predicates, however they do offer a straightforward mechanism for understanding the behavior and role of learned predicates. The ability to co-train GLIDR with other deep learning models also opens up a large design space in problems characterized by a mixture of symbolic and numerical data. Learned predicates could be combined with known ground truth predicates and learned rules can be combined with hard-coded known rules. In this way GLIDR and other differentiable rule learning systems can serve as a bridge between symbolic and numerical modalities.

\section{Strengths and Limitations}

\method\ learns expressive, graph-like rules that can't be represented by previous chain-like differentiable ILP techniques. This expressivity allows it to outperform other methods on knowledge graph completion tasks. The message passing inference algorithm required to support these expressive rules imposes the limitation that \method\ does not efficiently support open queries. 
This is in contrast with many embedding methods and chain-like rule methods such as Neural-LP \cite{yang2017differentiable} which can generate 1:N scores, emitting a confidence score for every tail entity given a head entity for the rule. While \method\ is still fast to train, the 1:1 limitation means that generating scores for knowledge graph completion requires $\mathcal{O}(|\mathcal{E}_{bg}|^2)$ inferences. Generating rankings for FB15k-237 took approximately 40 hours on 4 NVIDIA A100 80GB GPUs using the settings described. 

\section{Conclusion}
In this paper we introduce \method\ and show that its construction allows it to learn logical rules that cannot be represented by previous chain-like differentiable ILP methods. We evaluate the algorithm on several knowledge graph completion benchmarks and show that it achieves state of the art performance compared to other rule-based methods. We also perform experiments demonstrating that \method\ retains the characteristic noise robustness that differentiable ILP methods are known for and that hard rules extracted from GLIDR perform well. We also demonstrate the GLIDR can be co-trained with other deep learning models to incorporate rule-based learning in domains that are not inherently symbolic.
% Bibliography
\bibliographystyle{plain}
\bibliography{refs}

% Appendix
\appendix

\section{Experimental Details} \label{sec:hyperparams}
\subsection{Experiment Hyperparameters}
\begin{table}[htbp]
\centering
\caption{Hyperparameter settings used during Family, Kinships, and UMLS benchmarking.}
\label{tab:hyperparameters}
\begin{tabular}{ll}
\hline
\textbf{Hyperparameter} & \textbf{Value} \\
\hline
Training Steps         & 2048 \\
Batch Size             & 64 \\
Learning Rate          & 0.15 \\
Weight Decay           & 0.1 \\
Schematic Variables $N$ & 4 \\
Rule Bodies & 8 \\
Message Passing Rounds $R_{\text{max}}$ & 3 \\
\hline
\end{tabular}
\end{table}

We use the same hyperparameters across the Kinships, Family, and UMLS datasets, reported in Table \ref{tab:hyperparameters}. The learning rate and weight decay were chosen because we consistently observed these settings to contribute to stable convergence throughout GLIDR's development. The batch size was chosen because it offered acceptable training times across different GPUs. All sources of randomness including model weights and data-loaders were seeded with a different seeds across each of the 10 runs used to produce the statistics in Table \ref{tab:metrics}.
During GLIDR's development, we validated its performance extensively with the Hinton kinship dataset \cite{hinton1990kinship}, which has no noise. GLIDR classifies this dataset perfectly.

\begin{table}[htbp]
\centering
\caption{Hyperparameter settings used during FB15k-237 benchmarking.}
\label{tab:fb15k_hyperparams}
\begin{tabular}{ll}
\hline
\textbf{Hyperparameter} & \textbf{Value} \\
\hline
Training Steps         & 512 \\
Batch Size             & 64 \\
Learning Rate          & 0.15 \\
Weight Decay           & 0.1 \\
Schematic Variables $N$ & 4 \\
Rule Bodies & 8 \\
Training $R_{\text{max}}$ & 3 \\
Testing $R_{\text{max}}$ & 2 \\
\hline
\end{tabular}
\end{table}

Table \ref{tab:fb15k_hyperparams} records the hyperparameters used during benchmarking on the FB15k-237 dataset. A smaller number of training steps was chosen to reduce runtime, as was a reduced number of message passing steps at test-time.

\subsection{Hardware}
The experimental results reported in Table \ref{tab:metrics} were generated on a heterogeneous compute cluster, and so several different models of GPU and CPU were used throughout evaluation. Each of the benchmarking jobs had access to 8 CPU cores, 16GB of system memory, and one of the GPUs from the following list: NVIDIA RTX A6000-48GB, NVIDIA Tesla V100-32GB, NVIDIA Tesla V100S-32GB, NVIDIA A100-80GB. Training on YAGO3-10 \cite{dettmers2018conve}, a dataset significantly larger than any of those studied here, was validated on a NVIDIA RTX 3090-24GB GPU. Although training and evaluation would be impractically slow due to the large number of entities in this dataset (123,182), it demonstrates that memory complexity is not a significant concern for GLIDR given its use of sparse matrix algebra. Each benchmarking run on the Kinships, Family, and UMLS datasets took approximately 2 hours. The FB15k-237 benchmarking run used 4 NVIDIA A100-80GB GPUs, and the full run took approximately 72 hours, with approximately 48 hours of that time devoted to generating rankings.

\section{Examples of Learned Rules}
\begin{table*}[]
\begin{center}
\caption{A sample of learned rules for each studied dataset. Rules frequently contain terms resulting in non-chain-like branching or cyclic structure. All rules were fit with a maximum of $N=4$ schematic variables.}
\begin{tabular}{l l l}
\hline
\textbf{Dataset} & \textbf{Predicate} & \textbf{Learned Rule} \\
\hline
%$\texttt{result\_of}(X,Y):-$ & $\texttt{isa}(X,Z_1) \wedge \texttt{result\_of}(Z_1,Y).$ \\
UMLS & $\texttt{result\_of}(X,Y):-$ & $\texttt{result\_of}(X,Z_1) \wedge \texttt{co-occurs\_with}(Z_1,Z_2) \wedge $ \\
 && $ \texttt{result\_of}(Z_1,Y)\wedge\texttt{result\_of}(Z_2,Y).$ \\
&$\texttt{manifestation\_of}(X,Y):-$&$\texttt{exhibits}(Z_1,X)\wedge\texttt{disrupts}(Z_2,Z_1)\wedge\texttt{result\_of}(Z_1,Y)\wedge$ \\
&& $\texttt{complicates}(Z_2,Y).$ \\
&$\texttt{affects}(X,Y):-$ & $\texttt{affects}(X,Z_1)\wedge\texttt{interacts\_with}(X,Z_2)\wedge\texttt{affects}(Z_2,Z_1)\wedge$ \\
&& $\texttt{interacts\_with}(Z_1,Y)\wedge\texttt{affects}(Z_2,Y).$ \\
\hdashline
FB15k-237 & $\texttt{countries\_within}(X,Y):-$ & $\texttt{contains}(X,Z_1)\wedge\texttt{countries\_within}(X,Z_2)\wedge$ \\
&& $\texttt{adjoins}(Y,Z_2)\wedge\texttt{countries\_within}(Z_1,Y).$ \\
&$\texttt{place\_of\_burial}(X,Y):-$ & $\texttt{place\_of\_burial}(X,Z_1)\wedge\texttt{celebrity\_friendship}(X,Z_2)\wedge$ \\
&& $\texttt{place\_of\_burial}(Z_2,Z_1)\wedge$ \\
&& $\texttt{distributed\_films\_in\_region}(Z_2,Y).$ \\
& $\texttt{art\_direction\_by}(X,Y):-$ & $\texttt{set\_design\_for}(Z_1,X)\wedge \texttt{art\_direction\_by}(X,Z_2)\wedge$\\
&&$\texttt{award\_nomination\_for}(Y,Z_1)\wedge\texttt{award\_received\_for}(Y,Z_2).$ \\ \hdashline
Kinships & $\texttt{adardiya}(X,Y):-$ & $\texttt{anguriya}(Z_1,X)\wedge\texttt{adardiya}(X,Z_2)\wedge\texttt{adardiya}(Z_1,Y)\wedge$\\
&& $\texttt{anyainya}(Y,Z_2).$ \\
&$\texttt{agngiya}(X,Y):-$ & $\texttt{agngiya}(X,Z_1)\wedge\texttt{anguriya}(X,Z_2)\wedge\texttt{andungiya}(Z_2,Z_1)\wedge$ \\
&& $\texttt{awaadya}(Z_1,Y)\wedge\texttt{agngiya}(Z_2,Y).$ \\
& $\texttt{aleriya}(X,Y):-$ & $\texttt{adniadya}(Z_1,X)\wedge\texttt{aweniya}(Y,X)\wedge\texttt{umbaidya}(Y,Z_1).$ \\ \hdashline
Family & $\texttt{nephew}(X,Y):-$ & $\texttt{son}(X,Z_1)\wedge\texttt{daughter}(Z_2,Z_1)\wedge\texttt{uncle}(Y,Z_2).$\\
& $\texttt{aunt}(X,Y):-$ & $\texttt{sister}(X,Z_1)\wedge\texttt{sister}(X,Z_2)\wedge\texttt{brother}(Z_2,Z_1)\wedge$ \\
&& $\texttt{nephew}(Y,Z_1).$ \\
& $\texttt{father}(X,Y):-$ & $\texttt{husband}(X,Z_2)\wedge\texttt{mother}(Z_2,Y).$ \\ \hline

\end{tabular}
\label{tab:rules}
\end{center}
\end{table*}

Table \ref{tab:rules} reports three example rules learned for each dataset during benchmarking. Each reported rule performed well on the validation set compared to the other 7 competing rule definitions. Each of these rules was chosen because it illustrates interesting structure or describes a predicate for which the model performed particularly well. The predicates in the Kinships dataset are encoded as $\texttt{term1}, \texttt{term2}, $ etc., and have been translated to their corresponding Alyawarra kinship terms using the key provided by \cite{denham2016}. 

\end{document}